\documentclass[letterpaper, 10 pt, conference]{ieeeconf}  
\usepackage[colorlinks=true,linkcolor=blue,citecolor=blue,urlcolor=blue]{hyperref}

\IEEEoverridecommandlockouts                              

\usepackage{amsmath} 
\usepackage{graphicx}

\title{\LARGE \bf
Technical Report for ICRA 2025 GOOSE 2D Semantic Segmentation Challenge: Leveraging Color Shift Correction, RoPE-Swin Backbone, and Quantile-based Label Denoising Strategy for Robust Outdoor Scene Understanding
}

\author{
Chih-Chung Hsu, I-Hsuan Wu, Wen-Hai Tseng, Ching-Heng Cheng, Ming-Hsuan Wu, Jin-Hui Jiang, Yu-Jou Hsiao \\
Institute of Intelligent Systems College of Artificial Intelligence, National Yang Ming Chiao Tung University \\
Institute of Data Science, National Cheng Kung University
}

\usepackage{fancyhdr}
\fancypagestyle{withfooter}{

\fancyfoot[C]{\footnotesize Winners of the GOOSE 2D Semantic Segmentation Challenge at the IEEE ICRA Workshop on Field Robotics 2025}
}

\begin{document}

\maketitle
\thispagestyle{withfooter}
\pagestyle{withfooter}

\begin{abstract}

This report presents our semantic segmentation framework developed by team ACVLAB for the ICRA 2025 GOOSE 2D Semantic Segmentation Challenge, which focuses on parsing outdoor scenes into nine semantic categories under real-world conditions. Our method integrates a Swin Transformer backbone enhanced with Rotary Position Embedding (RoPE) for improved spatial generalization, alongside a Color Shift Estimation-and-Correction module designed to compensate for illumination inconsistencies in natural environments. To further improve training stability, we adopt a quantile-based denoising strategy that downweights the top 2.5\% of highest-error pixels, treating them as noise and suppressing their influence during optimization. Evaluated on the official GOOSE test set, our approach achieved a mean Intersection over Union (mIoU) of 0.848, demonstrating the effectiveness of combining color correction, positional encoding, and error-aware denoising in robust semantic segmentation.

\end{abstract}

\section{INTRODUCTION}

The ICRA 2025 GOOSE\cite{mortimer2023goosedatasetperceptionunstructured} 2D Semantic Segmentation Challenge aims to benchmark the robustness of segmentation models deployed in complex, real-world robotic scenarios. The task involves assigning one of nine semantic categories to each pixel in outdoor environments captured by four heterogeneous robotic platforms—ALICE, MuCAR-3, Spot v1, and Spot v2—each operating with different camera specifications, sensor viewpoints, and scene dynamics. The dataset includes high-resolution images ranging from 1280×720px to 2048×1536px and spans a wide variety of terrains and lighting conditions.

A key challenge posed by this dataset lies in its cross-platform variability. Models must not only perform well on individual platforms but also generalize across robots with differing viewpoints, illumination profiles, and camera resolutions. Compounding this is the presence of annotation noise, which is common in large-scale, manually labeled datasets—especially those collected in uncontrolled outdoor environments. Mislabeling, ambiguous boundaries, and inconsistent class definitions can degrade the reliability of learned representations and destabilize training. These difficulties are further reflected in the evaluation protocol: the mean Intersection over Union (mIoU) is computed separately for each robot, ignoring the "other" class, and aggregated using a weighted average—67\% from MuCAR-3, 24\% from ALICE, 6\% from Spot v2, and 3\% from Spot v1—based on test split proportions.

To tackle the challenges arising from domain shifts and annotation noise in outdoor scene understanding, we propose a segmentation framework based on MaskDINO\cite{li2022maskdinounifiedtransformerbased}, a Transformer-based image segmentation model that has recently surpassed conventional CNN-based approaches—such as UNet\cite{ronneberger2015unetconvolutionalnetworksbiomedical}, DeepLabV3+\cite{chen2018encoderdecoderatrousseparableconvolution}, and PSPNet\cite{zhao2017pyramidsceneparsingnetwork}—in both accuracy and generalization. Transformer architectures such as MaskDINO leverage global attention and dense contextual modeling, enabling superior performance on dense prediction tasks. These strengths make MaskDINO particularly well-suited for the diverse and visually complex outdoor environments targeted in this work.

Before settling on MaskDINO as the backbone of our framework, we also evaluated other state-of-the-art Transformer-based segmentation models, including Mask2Former\cite{cheng2021mask2former} and PEM\cite{cavagnero2024pem}. While each demonstrated competitive results, MaskDINO consistently outperformed them in terms of segmentation accuracy and robustness across domains, ultimately making it the most suitable foundation for our task.

To further enhance performance under outdoor scene variability, we integrate three key components into our MaskDINO-based pipeline. First, a Color Shift Estimation-and-Correction (CSEC) module \cite{li_2024_cvpr_csec} is employed to address the color tone distortions caused by inconsistent natural lighting, effectively normalizing illumination across scenes. In addition to color shifts, low-light or shadowed regions in outdoor environments often degrade segmentation performance due to suppressed texture and contrast information. By correcting illumination artifacts across both overexposed and underexposed areas, the CSEC module helps stabilize the visual representation and improves feature reliability for subsequent processing.

Second, we upgrade the backbone with a RoPE\cite{heo2024ropevit} enhanced variant of the Swin Transformer\cite{liu2021Swin},incorporating rotary positional embeddings that enhance spatial generalization, particularly under resolution and scale variations common in outdoor imagery. Lastly, to mitigate the adverse effects of noisy labels, we introduce a quantile-based denoising mechanism that identifies and downweights the top 2.5\% of high-error pixels during training, treating them as unreliable and reducing their impact on gradient updates.
\begin{figure*}[htbp]
    \centering
    \includegraphics[width=\textwidth]{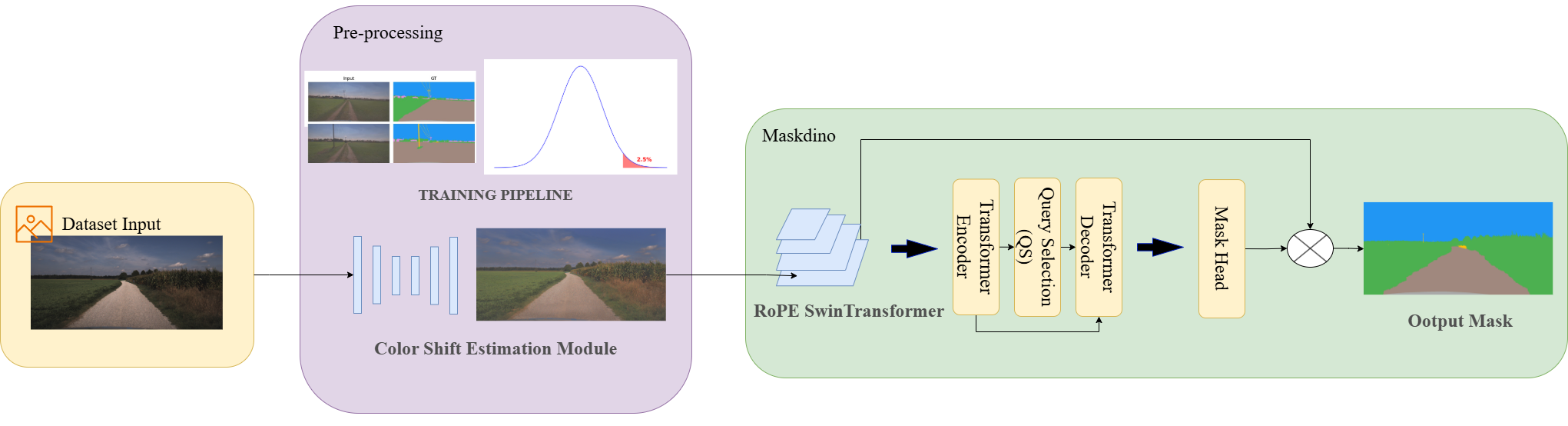}
    \caption{Architecture of Robust Outdoor Scene Understanding with Color Shift Correction, RoPE-Swin Backbone, and Quantile-based Denoising}
    \label{fig:model-architecture}
\end{figure*}

Together, these enhancements allow our framework to achieve greater robustness and accuracy, particularly in the presence of challenging lighting conditions, spatial distortions, and annotation imperfections.

These integrated techniques allow our model to maintain segmentation consistency across platforms and mitigate the negative effects of label noise, achieving a competitive mIoU of 0.848 on the official GOOSE test set. Our results highlight the importance of combining transformer-based architectures with robustness-oriented training strategies for real-world robotic perception tasks.

\noindent\textbf{Our contributions are summarized as follows:}
\begin{itemize}
    \item We propose a robust segmentation framework based on MaskDINO~\cite{li2022maskdinounifiedtransformerbased}, which outperforms other Transformer-based models in outdoor scene understanding tasks.
    \item We integrate a Color Shift Estimation-and-Correction (CSEC) module~\cite{li_2024_cvpr_csec} to address illumination inconsistency and enhance visual stability under varying lighting conditions.
    \item We enhance the Transformer backbone using RoPE~\cite{heo2024ropevit} to improve spatial generalization across scales and resolutions.
    \item We introduce a quantile-based label denoising strategy that mitigates the impact of annotation noise by downweighting unreliable supervision during training.
    \item Our full pipeline achieves a competitive mIoU of 0.848 on the GOOSE test set, demonstrating robustness to color shifts, spatial distortions, and label noise in real-world outdoor scenes.
\end{itemize}

\section{Method}

Our method aims to enhance semantic segmentation performance in outdoor scenes, particularly under challenging conditions such as significant lighting variations and unstable image quality. As illustrated in Figure \ref{fig:model-architecture}, our approach builds upon the MaskDINO architecture with three key improvements: (1) integration of the Color Shift Estimation-and-Correction (CSEC) model proposed by Yiyu Li et al. for image correction, (2) replacement of the original backbone with the RoPE-ViT Transformer incorporating Rotary Positional Embedding as proposed by Byeongho Heo et al., and (3) implementation of a training data filtering strategy to exclude low-quality samples.

\subsection{Color Shift Estimation-and-Correction (CSEC) Model}

In practical applications, we observe that a significant portion of training images suffer from overexposure or underexposure, leading to severe brightness and color shifts in object regions. This adversely affects the accuracy of semantic region discrimination and mask prediction. To mitigate the inconsistencies caused by such image quality issues, we incorporate the CSEC model during the data preprocessing stage. The CSEC model comprises two key modules: the Color Shift Estimation (COSE) module and the Color Modulation (COMO) module.

\subsubsection{Color Shift Estimation (COSE) Module}

The COSE module estimates the color shift present in the input image, which arises due to uneven lighting and other factors causing global and local color anomalies. By employing a deep learning-based offset prediction mechanism, it effectively detects and quantifies color deviations in the image, providing a basis for subsequent correction operations. Specifically, the COSE module operates as follows:

\begin{equation}
y = \sum_{i=1}^{N} w_i \cdot x_i + \Delta p_i
\end{equation}

where $x_i$ represents the input feature maps, $w_i$ denotes the convolution kernel weights, $\Delta p_i$ is the spatial offset, and $y$ is the output feature map.

\subsubsection{Color Modulation (COMO) Module}

Based on the offset information output by the COSE module, the COMO module maps the original image from an abnormal color space back to a natural and balanced representation. Specifically, the COMO module utilizes the darkening offset $\Delta d$ and brightening offset $\Delta b$ generated by the COSE module to adjust the color and brightness, restoring image details and improving issues of overexposure or underexposure. Assuming $X$ is the input image, and we have obtained the corresponding darkening offset $\Delta d$ and brightening offset $\Delta b$, the COMO module operates as follows:

First, we extract the feature map $F_X$ from the original image $X$, and then extract features from the darkening and brightening offsets $\Delta d$ and $\Delta b$, obtaining $F_d$ and $F_b$, respectively. Next, we compute the self-correlation matrices $A_X$, $A_d$, and $A_b$, and fuse these matrices with learned weights to adjust the image's color and brightness. The final corrected feature map $F_{\text{corr}}$ is expressed as:

\begin{align}
F_{\text{corr}} =\; & \gamma_X \cdot \text{SymNorm}(A_X) \cdot F_X \nonumber \\
& + \gamma_d \cdot \text{SymNorm}(A_d) \cdot F_d \nonumber \\
& + \gamma_b \cdot \text{SymNorm}(A_b) \cdot F_b + b
\end{align}

where $\gamma_X$, $\gamma_d$, and $\gamma_b$ are the learned fusion weights, $\text{SymNorm}(A) = D^{-1/2} \left( \frac{2A + A^\top}{2} \right) D^{-1/2}$ represents the symmetrization and normalization operation, $D$ is the diagonal matrix of $\frac{2A + A^\top}{2}$, and $b$ is the bias term.

The final corrected image $I_{\text{corr}}$ is reconstructed by the decoder module:

\begin{equation}
I_{\text{corr}} = \text{Decoder}(F_{\text{corr}})
\end{equation}

\subsection{RoPE-ViT Backbone Integration}

To enhance global semantic modeling capabilities, we replace the default Swin-L backbone in MaskDINO with RoPE-ViT. This model is based on the Vision Transformer architecture and incorporates Rotary Positional Embedding (RoPE), which improves the modeling of relationships between image patches without increasing computational costs, facilitating subsequent semantic segmentation tasks.

RoPE applies position-related rotational transformations to the Query and Key vectors in the self-attention mechanism, directly integrating relative positional information into the attention's inner product computation. Specifically, RoPE mimics the design of sinusoidal encoding \cite{vaswani2023attentionneed}, splitting each embedding vector into multiple 2D sub-vectors and applying the following rotation operation to each pair of dimensions:

\begin{equation}
\text{RoPE}(x_{2i}, x_{2i+1}) = 
\begin{bmatrix}
x_{2i} \cos(\theta_i) - x_{2i+1} \sin(\theta_i) \\
x_{2i} \sin(\theta_i) + x_{2i+1} \cos(\theta_i)
\end{bmatrix}
\label{eq:rope}
\end{equation}

where $\theta_i = p \cdot \omega_i$,  
$p$ represents the relative position index of the image patch, and  
$\omega_i = 10000^{-2i/d}$ corresponds to the frequency for each dimension.

\subsection{Quantile-based Label Denoising Strategy}

Considering the presence of annotation errors or anomalous samples that are difficult to learn in real-world data, we design a data filtering strategy during the training process. Specifically, we use the existing model to predict each training sample and calculate the pixel-wise error rate between the predicted mask and the corresponding ground truth mask. After statistically analyzing the error rates of all training samples, we remove samples that fall above the 97.5th percentile of the error rate distribution, retaining the relatively normal samples below the 97.5th percentile. This strategy helps eliminate extreme outliers, enhancing the model's learning stability and generalization capability.

\section{Experiments}

To validate the effectiveness of our proposed approach, we conduct extensive experiments on the GOOSE dataset\cite{mortimer2023goosedatasetperceptionunstructured}. We evaluate model performance using the commonly adopted metric, mean Intersection-over-Union (mIoU). During training, we set the batch size to 4 and utilize two NVIDIA RTX 4090 GPUs. In the following sections, we first benchmark our method against existing approaches and then perform a series of ablation studies to demonstrate its effectiveness and generalization capability.

\begin{table}[ht]
\centering
\begin{tabular}{l c}
\hline
Method & mIOU (\%) \\
\hline
(w/o) RoPE & 88.18 \\
(w/o) CSEC & 88.72 \\
(w/o) RoPE \& CSEC & 87.92 \\
MaskDINO + RoPE + CSEC & \textbf{88.89} \\
\hline
\end{tabular}
\caption{mIoU Metric Comparison of Different Methods on the GOOSE Validation Set.}
\label{tab:exp_results}
\end{table}

\begin{table}[ht]
\centering
\begin{tabular}{l c c}
\hline
Method & Denoise & mIOU (\%) \\
\hline
MaskDINO + RoPE + CSEC & x & 88.89 \\
MaskDINO + RoPE + CSEC & v & \textbf{89.13} \\
\hline
\end{tabular}
\caption{mIoU Metric Comparison with and without Denoising on the GOOSE Validation Set.}
\label{tab:exp_results_denoise}
\end{table}

We systematically evaluated the impact of the two key components in our proposed framework: (1) replacing the MaskDINO backbone with a variant incorporating Rotary Position Embedding (RoPE), and (2) introducing the CSEC image enhancement strategy. Table~\ref{tab:exp_results} summarizes the performance of four configurations: the original MaskDINO, MaskDINO with RoPE backbone only, MaskDINO with CSEC enhancement only, and the full model combining both components. All evaluations were conducted on the validation set.

The results show that the model integrating both RoPE and CSEC achieves an mIoU of 88.89\%, outperforming all other settings. Introducing either RoPE or CSEC individually also yields performance gains, validating the positive contribution of both modules.

\begin{figure}[htbp]
    \centering
    \includegraphics[width=0.95\linewidth]{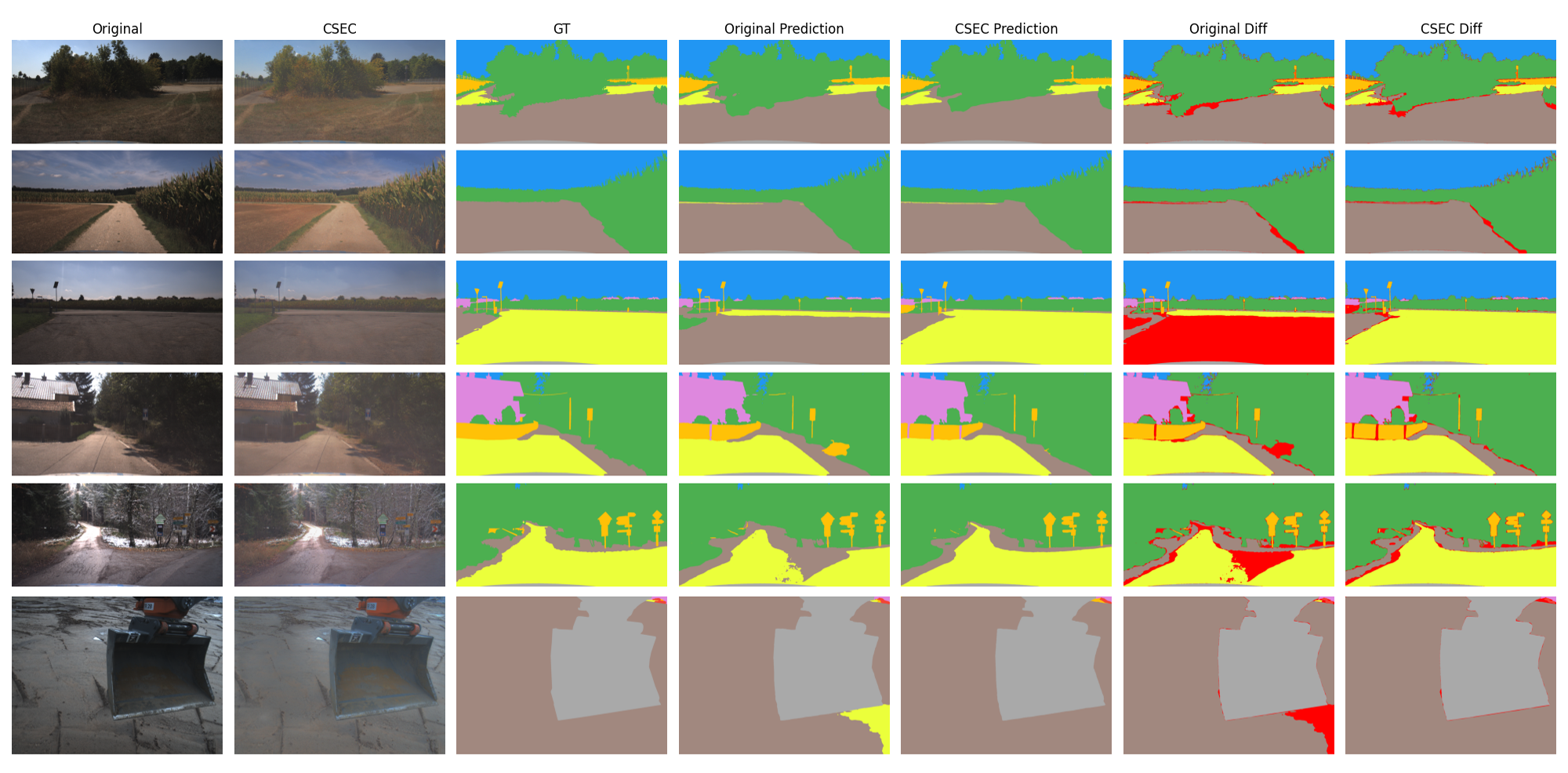}
    \caption{Comparison of segmentation results between CSEC-enhanced model and baseline}
    \label{fig:csec_vis}
\end{figure}

\begin{figure}[htbp]
    \centering
    \includegraphics[width=0.95\linewidth]{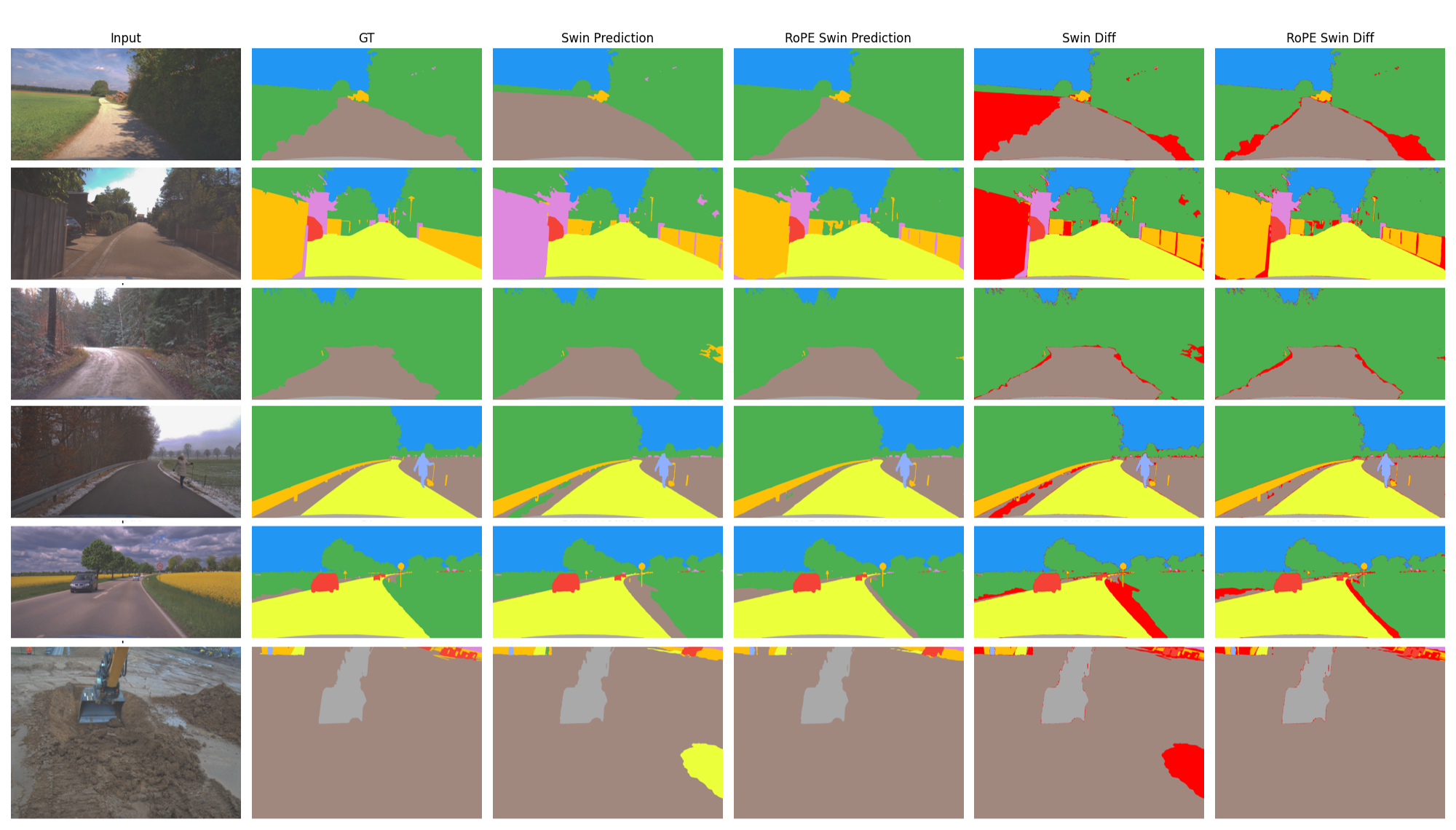}
    \caption{Comparison of segmentation results with original and RoPE backbone}
    \label{fig:rope_vis}
\end{figure}

Figures~\ref{fig:csec_vis} and~\ref{fig:rope_vis} provide qualitative visualizations of the results. Figure~\ref{fig:csec_vis} demonstrates that the model with CSEC produces more accurate segmentations along object boundaries and fine details, particularly in scenarios with uneven color distribution where missegmentation is common. Figure~\ref{fig:rope_vis} shows that although modifying the backbone does not substantially alter segmentation contours, it improves classification accuracy and semantic consistency.

Building upon the previous results, we further investigated the effect of the Training Data Filtering Strategy. Table~\ref{tab:exp_results_denoise} compares model performance with and without data filtering under the combined RoPE and CSEC setting. By filtering out anomalous samples to reduce dataset noise, the model's performance improved to an mIoU of 89.13\%, indicating that this strategy effectively enhances training data quality and segmentation capability.

It is important to note that all aforementioned evaluations were conducted on the validation set. Our final model—integrating the RoPE backbone, CSEC enhancement, and the data filtering strategy—achieves an mIoU of 84.8\% on the test set, demonstrating strong generalization ability and robustness.

\section{CONCLUSIONS}

In this work, we have presented a robust semantic segmentation framework designed for complex outdoor environments, leveraging the strengths of MaskDINO and introducing three key enhancements—Color Shift Estimation-and-Correction (CSEC), a RoPE-ViT backbone, and a quantile-based label denoising strategy. Our method effectively addresses the challenges of domain shifts, lighting variations, and annotation noise that commonly degrade segmentation performance in real-world robotic applications.

Extensive experiments on the GOOSE dataset demonstrate that our approach outperforms existing Transformer-based models, achieving a state-of-the-art mIoU of 84.8\% on the test set. Specifically, our ablation studies highlight the importance of each proposed component: CSEC significantly enhances image quality under diverse lighting conditions, RoPE-ViT improves spatial generalization, and quantile-based denoising mitigates the impact of unreliable labels. These improvements enable our model to maintain high segmentation accuracy across different robotic platforms and environmental conditions.

\addtolength{\textheight}{-12cm}   








References are important to the reader; therefore, each citation must be complete and correct. If at all possible, references should be commonly available publications.


\bibliographystyle{IEEEtran} 
\bibliography{main} 

\end{document}